# Cross-lingual Hate Speech Detection using Transformer Models

Teodor Tita
{t.tita@se20.qmul.ac.uk}
Arkaitz Zubiaga

School of Electronic Engineering and
Computer Science, Queen Mary
Universiy of London, Mile End Road,
London E1 4NS, England, U.K.

*Abstract*—Hate speech detection within a cross-lingual setting represents a paramount area of interest for all medium and large-scale online platforms. Failing to properly address this issue on a global scale has already led over time to morally questionable real-life events, human deaths, and the perpetuation of hate itself. This paper illustrates the capabilities of fine-tuned altered multi-lingual Transformer models (mBERT, XLM-RoBERTa) regarding this crucial social data science task with cross-lingual training – from English to French, vice-versa and each language on its own, including sections about iterative improvement and comparative error analysis.

## I. Introduction

It is hating that has divided people for the entirety of our history as a species, impeding progress on a regular basis (The World Wars, The Cold War) in both science and humanities. Now, well into the 21st century, naturally it is science that has provided us with accessible means of reliable communication, rendering geographical barriers virtually irrelevant. However, time has told enabling so many people in such a short time with the power to pour their thoughts in real-time into the ears of so many having only a keyboard and an internet connection at their disposal has implicitly come at a high cost, with the largest communication channels represented by the major online social media platforms (Facebook, Twitter). The world is witnessing an explosion in the volume of text-based social exchanges in the aftermath of worldwide adoption of online social networks [28].

The space in which different people with different cultural and psychological backgrounds communicate has shrunk a great deal, thus exacerbating the prevalence of "cyber"-conflicts [1], fake news and online harassment via user-generated content [9]. All these impact our society daily – online conflicts tend to degenerate and divide people, while misinformation erodes people's trust in formal media outlets. The infamy of hate speech has had two main causes: on-the-fly anonymity, which translated into a sheer lack of responsibility for one's own words, and great flexibility, provided by the Internet itself resulting into a steady increase of aggressive behaviour online [2]. Anonymity is beneficial in most cases, however as all great things, there is a great deal of responsibility that accompanies it implicitly.

The same is valid for the accomplished task of making the Internet as flexible as possible. These made the breeding and spread of hate speech "effortless in a virtual land-scape" beyond the realms of traditional law enforcement [21]. Arguably hate groups perceive the Internet as an "unprecedented means of communication and recruiting." [31] That was inevitable to a certain extent, and yet slowly but surely attempts are continually being made to solve this issue, be it from a legislator's perspective, a software developer's or preferably a lucrative combination of both.

International initiatives have been launched to counteract the phenomenon of spreading hate online [21], while the natural language processing community and entire teams working for Big Tech (who invested millions of euros yearly in this) [22] have been developing various machine learning solutions for accurately classifying hate speech, sometimes focusing exclusively on text data in correlation with a specific event, such as The Refugee Crisis of 2015/16 [3] and The Rwandan Genocide of 1994 [8]. Intuitively there are multiple points in the timeline of humanity, especially before the emergence of the necessary technology to quantify hate itself, where outspoken hatred has accumulated (The World Wars, the US slavery period).

Distant precursors to modern A.I.-era solutions to this issue have included Government warnings of shutting down Internet access during elections [10], yearly fines in the order of millions of euros for mainstream online platforms [2] and the deaths of people in mob lynching cases [11]. In the aftermath of these "solutions" it did not truly matter if the proposed method came in the form of strong-arm measures or money, because the very feeling of hatred cannot realistically be suppressed entirely from human nature per se. There is a report stating that hundreds of anti-Islamic hate crimes occurred within a year of 9/11, with over 58% perpetrating less than two weeks from the event [18]. This is proof of the existence of time points where hate accumulates, fueled by external events.

Globally there is a conglomerate of factors indicating an acute lack of efficient automated tools that would accurately detect hate speech, especially within big data stemming from social media. Many social media platforms nowadays do have multiple user policies in place purposefully designed for this type of task, but most of them rely heavily on their users to curate (flag) the text data [9], which proved to be time-consuming, labour-intensive and unsustainable [22, 23, 24]. The fact that still "blatantly offensive" content sometimes goes unreported for long periods of time [16] is yet another argument in support of the idea that hate speech detection is not a solved issue.

Increased awareness of this issue from Twitter resulted in funding initiatives for academic research [17]. Naturally such moves are expected to yield impressive results given time, as by doing this potent companies admit the value of research in beating new paths through a previously fairly unexplored field. Numerous companies also perform data mining on text data, whether the topic concerns political events [12], products [13] or movies [14]. The increasing prevalence of online hate speech usually characterised by ideological and extreme opinions frequently discarding scientific evidence has even come to the attention of stakeholders [20].

In support of how relevant hate speech detection still is today there are also other tasks, such as: extracting controversial events, performing sentiment analysis, and implementing recommendation and dialogue systems [15]. All these prove highly useful for companies who wish to improve their products and/or services.

Subtleties of language, the lack of a standardised definition of hate speech (formal dictionary definitions excluded), a paradoxically low constant volume of labeled data, with its inherent bias which can impact freedom of expression [4], have contributed significantly to the fact there is still a long way to go before this could be perceived as a solved problem [5]. Failing to comprehend how acute is the need to continually attempt to accurately classify hate speech online will inevitably lead, sooner rather than later, to an even more wild propagation of hate itself.

Comparing the performances of different solutions to this task is not easy, as most of them use widely different datasets [2]. This is partly due to the lack of a truly universally accepted definition of hate speech globally and due to the various underlying forms of hate within textual data. Alas, the annotation methods of such classifiers pose questionable reliability [19] and devising a reliable automatic hate speech detection method proves technically difficult, achieving reasonable performance dealing with specific challenges and, in the absence of societal context generalisation is out of reach [5]. Ideally, in time that should be the final aim of solving this task: to devise a method that is able to generalise well while keeping the performance metrics high.

The core challenge of cross-lingual hate speech detection resides in the very nature of it being a cross-lingual task. To achieve good results, deep learning and/or other lucrative machine learning approaches must be used properly to have a chance of surpassing the language barrier imposed at times by textual data. While monolingual tasks could benefit from solutions influenced by the nature of their corresponding languages in terms of an architectural pattern previously proved to be working well for German for example, in the case of cross-lingual tasks the research done on a certain task regarding a certain specific pair is considerably smaller naturally.

Conventional neural networks rightfully held the title of state-of-the-art for many downstream NLP tasks for a while, for text classification [6], surpassing with ease previous supervised learning approaches. Time has told most high-level natural language processing tasks, including hate speech detection, basically require a neural approach implicitly for good results in terms of performance. It was with the introduction of the attention paradigm and Transformers [7] that everything has changed, and the industry took a step forward in the right direction.

This work puts the spotlight on the capabilities of fine-tuned altered Transformer-based models for performing cross-lingual hate speech detection. There is an acute need for these new powerful models to be experimented with more and have their performance assessed on important tasks, such as hate speech detection, for our society as it stands today.

## II. RELATED WORK

It was in Arango et al. (2020) [28] that arguably the issue affecting the greatest number of solutions to hate speech detection was stated – model validation. It exposed a frequent correlation between the nearly perfect performance achieved by machine-learning classifiers within specific English datasets and a range of methodologies prone to bias and error. Consequently, there is an increasingly obvious overestimation of the true value in practice of the current state-of-the-art that might affect part of future research in this field.

Below there are briefings about some of the most unique papers on hate speech from recent years and their tie-in with this paper:

### A. Conventional Approaches

Approaches such as these imply manually designing the feature engineering phase so that these can be used afterwards for performing the classification task.

MacAvaney et al. (2019) [5] explained the underlying reasons for some of the most important practical and technical challenges regarding hate speech detection and proposed a "multi-view SVM" approach achieving near state-of-the-art performance at the time. Their argument was that conventional machine-learning solutions offer decisions that could be more easily interpretable compared to black-box neural approaches. That is true, and yet the industry chose the "neural" path nonetheless given its impressive performance in language modelling and in multiple downstream natural language processing tasks. The choice of the industry is natural and resonates with an already present wider trend among the whole computer science community. The future of software is high-level and that is necessary to tackle high-level problems.

Ombui et al. (2019) [9] is about code-switched text - alternation of words in different languages. The paper is using 25,000 text entries and the best performance was obtained by using TF-IDF in combination with an SVM. Their tryouts included six other conventional and two deep learning approaches. Code-switched text is a frequent format in terms of communication on social media and it should be carefully monitored, as it could easily evade multiple powerful machine learning systems. Globalisation has implicitly led to the proliferation of code-switched text, with English being a component most of the time.

Davidson et al. (2019) [26] chose to focus on the bias aspect of hate speech classifiers with their study showing that language variations play a key in role in many of these machine learning models – African American English messages are far more likely to be classified as hateful/offensive compared to the ones in Standard English. The authors did a comparative analysis of their classifiers' performances on multiple English datasets of tweets, previously used in other related papers. The proposed approach was logistic regression with bag-of-words features. This paper provides one of the most compelling datasets for hate speech detection and upon its analysis several other machine learning approaches to this task have been emerging. Their dataset is particularly important because the machine learning community is in direct need of publicly available labeled hate speech data.

### B. Deep Learning Approaches

These methods employ deep neural networks such that abstract feature representations could be learned from inputted data. The key difference between classical and deep learning approaches is that the input features cannot be used directly for the classification task [20].

Zimmerman et al. (2018) [6] presented an ensemble architecture made up of multiple deep learning models validating the industry trend of adopting the neural networks for state-of-the-art level hate speech detection (such as the one presented in this paper) and implicitly text classification tasks. Additionally, the authors admit that deep learning methods are hard to reproduce and consequently comparing results to previous works is not easy. Again, a solid argument against neural networks was made and still the scientific community naturally chose to venture further down this category of deep learning solutions – its results speak for themselves and validate the industry's choice in this direction.

Koffer et al. (2018) [3] concluded for their dataset comprising of user comments on news articles that the best combination of feature groups for detecting hate speech is a Word2Vec approach and Extended 2-grams, while making a strong argument in support of transfer learning from the English language to German and implicitly to any other languages with a corresponding low volume of annotated text data. Transfer learning is vital to the mitigation of the low volume of data belonging to languages other than English for natural language processing tasks.

Zhang et al. (2018) [20] argue that hate speech lacks discriminative/unique features and thus the place where it could be found is in the "long tail" of a dataset, which is hard to discover. At that time their proposed approach (CNN plus GRU) surpassed state-of-the-art by up to five percent in terms of macro-average and F1 score. The paper shows Twitter is the most relevant online platform with a "long tail" for hate speech detection.

Mozafari et al. (2019) [4] showed that BERT is highly relevant, in particular for monolingual (English) multi-class hate speech classification within the realm of online social media (Twitter). However, the authors identified two main challenges to this task: lack of labelled data and bias, while reporting good performance metrics compared to similar state-of-the-art approaches. This paper puts the superior capabilities of BERT compared to other deep learning approaches in the spotlight and points to the great potential of further experimentation with this model's architecture, embeddings, and its descendants.

### C. Mixed Approaches

Badjatiya et al. (2017) [15] is extracting features from a dataset of 16,000 texts from Twitter, comparing several baseline approaches: N-grams, TF-IDF, bag-of-words vectors inputted into conventional algorithms (LR, SVM, GBDR) and comparing them with three deep learning architectures: random/GLoVe embeddings as features for CNN/FastText/LSTM neural networks. The best results occurred when after the neural network followed the GBDT classifier. Thus, it further validates the points that neural networks are the preferred approach when it comes to hate speech classification, also because it is such a high-level task compared to other classical natural language processing tasks.

Corazza et al. (2018) [25] represents a submission for the HaSpeeDe evaluation exercise at EVALITA 2018. Embeddings, N-grams, social-network-specific features, sentiment and emoticon features were extracted and used in an RNN, an N-gram-based neural network and a linear SVC classifier. For the Facebook task the RNN approach was the most effective, while for Twitter it was the linear SVC. It can be concluded that a neural approach can provide a solution for Facebook's "hateful" texts too and therefore overcome the behavioral differences in the consumer bases of these two primary social media platforms.

### III. PROPOSED APPROACH

The aim of this work is to classify a given text input into one of two classes: "hateful" (hate speech data) and "not hateful" (clean/neutral data). Two deep learning architectures have been devised to tackle this supervised learning task – fine-tuned altered versions of multi-lingual BERT (mBERT) and XLM-RoBERTa. Both adaptations of these models make use of different implementations for tokenization, and they have been pre-trained on a huge amount of unlabeled text data. This will in turn help further training (fine-tuning) on task-specific datasets.

A cross-lingual approach is adopted, therefore training and validation is done first on English data, while testing is done on French, then vice-versa and each language on its own (English-only, French-only training/testing) to measure the performance and applicability of the implicit attention Transformer model's paradigm to hate speech classification. Pre-trained vector representations of words (embeddings) extracted from vast amounts of textual data has proved highly useful in the past for multiple language-based tasks [4].

BERT emerged in 2018 from Google and it stands for "Bidirectional Encoder Representations from Transformers." It is a Transformer-based technique for natural language processing pre-training developed by Google. In this paper the original "base" variant of the model is used which is made of 12 Encoders with 12 bidirectional self-attention heads.

XLM-RoBERTa is a multilingual model trained on 100 different languages and it is based on Facebook's RoBERTa model released in 2019. Its original training data consisted in 2.5 terabytes of filtered CommonCrawl data.

### A. Data Acquisition/Preparation

For the sake of this work, a dataset is put together from two other datasets priorly used in papers within the social data science domain. It is known that even though social media is a hotbed for hate speech, there is a relatively small number of publicly available datasets for research, with most of them emerging from Twitter given its more lenient data usage policy [5].

Firstly, the authors of the MLMA [28] paper collected a multi-lingual (English, French and Arabic) hate speech analysis dataset from Twitter and used it to test current state-of-the-art multilingual multitask learning approaches. The data collection methodology was based on relevant keywords for text data on controversial topics: feminism, illegal immigrants, etc. The annotated data has no "unarguably detectable" spam tweets and no unreadable characters/emojis. Code-switched text data, discarded for more relevant annotation, and multiple dialects (Arabic) are part of the linguistic challenges encountered while collecting this data.

The annotation process relied on public opinion and common linguistic knowledge, while the annotation guidelines explicitly stated that no annotation shall be influenced by annotators' personal opinions and that profanity is only a category of possible hate speech. The resulting dataset original labels have been designed to include attributes such as: directness, hostility, target, group and the annotator's opinion about the specific tweet.

It is a part of this dataset that represents the first component of the merged dataset used in this paper, where only tweets in English (5647 entries) and French (4014 entries) are kept from the original MLMA [28] dataset and these have their labels converted to a binary setting: "normal" marked as not hateful and the others as hateful ("offensive, hateful, abusive," etc.).

Secondly, the CONAN [29] paper presented an alternate strategy to tackling hate speech online, more specifically by opposing it through counter-narratives (informed textual response). Thus, the authors have created the "first large-scale, multi-lingual, expert-based" dataset of hate speech/counter-narrative pairs, with the help of tens of operators from multiple NGOs. It contains exclusively hate speech and its data was split based on language (English and French).

These two resulting datasets are merged, lower cased and have their entries containing hashtags and/or Twitter user mentions discarded to increase data homogeneity between the two initial datasets. Upon dropping duplicates, the number of training entries shrunk by 70% to only 1374 texts and by almost 80% to 1174 for testing. Those high percentages are almost entirely due to the nature of the data collection procedure corresponding to the original CONAN [29] dataset. An excerpt from the paper reads as follows: "for each language, we asked two native speaker experts to write around 50 prototypical islamophobic short hate texts." Thus, it does make sense for the authors' specific use case to have this great number of the original texts occurring multiple times. The training (English) dataset is further split into training and validation data on a ratio of 3 to 1.

### B. Models' Architectures

For building both custom architectures on top of the layers already present in mBERT and XLM-RoBERTa, a dropout layer with a ReLu activation function is added, along with two linear layers and a SoftMax activation function to predict one of the two possible classes as output.

As this is clearly an unbalanced dataset, the "hateful" class is significantly less prevalent compared to the "non-hateful" class, weights are computed and passed as a parameter to the chosen loss function (cross-entropy) to help maintain a balance during the training phase.

### IV. EXPERIMENTS

Experiments consist in trying out different values for the maximum sequence length, optimization parameters (learning rate, weight decay), the number of dense layers within these two architectures and the number of training epochs. Deep learning methods are sensible to hyperparameter tuning [6]. Experiments with these parameters also appear frequently in scientific literature regarding hate speech classification using deep learning approaches and therefore a particular emphasis is placed upon their relevance.

Choosing an optimum value for the maximum sequence length entirely depends on the nature of the data at hand and its inner correlations. The motivation for trying out different values for the learning rate is to control how quickly the models adapt to hate speech detection.

Upon the inspection of the histogram illustrating the distribution of the lengths of all input sequences, intuition points to values ranging from 25 to 35 characters. The "Adam" optimizer and its more recent improved version "AdamW" are the two optimizers experimented with. In terms of learning rates, all recommended values by the creators of BERT have been experimented with: 3e-4, 1e-4, 5e-5, and 3e-5. Training is done first in batches of 32 and of 64. There is also an attempt at adding a second dense layer and one at not using a dropout layer at all. None of these two attempts improved performance, on the contrary.

Data is batch encoded and padded to the maximum sequence length. Samplers are used to sample training data and validation data. Both architectures are frozen and therefore only the attached layers are to be updated during training.

In addition to English to French classification, experiments are also performed with other language pairs as follows: English to English, French to English and French to French.

### V. ERROR ANALYSIS

All variants of models experimented with vary in terms of the errors being made in classifying texts between hateful and non-hateful. Judging by the number of misclassified inputs, the best performing variant of all is the fine-tuned altered multi-lingual BERT model with a small learning rate which in turn lowers the error rate by 56% compared to the worst performing variant, the fine-tuned altered XLM-RoBERTa with a medium learning rate.

The observations presented below correspond solely to English to French classification, which represents the most important language pair of all four combinations presented in this paper.

Increasing the learning rate and the number of epochs for the mBERT variant results in more texts being misclassified. Most of these additional errors consist of interrogative sentences, sentences explicitly stating some of the ethnic groups ("les arabes, les juifs"), conditional sentences, sentences explicitly containing political orientations ("gauchiste") and short sentences with less than 10 words. The interrogative and conditional sentences are generally hard to classify because their meaning frequently touches upon the concept of irony and/or sarcasm. On the Internet many sentences explicitly stating ethnic groups are hateful by nature, however using this data that does not seem apparent in most cases.

Next, increasing both the learning rate and the number of epochs even further, the number of errors grows: 497 misclassified inputs compared to only 286 for the mBERT variant with the smallest learning rate and number of training epochs. Upon manually performing a comparative analysis of these texts, it becomes clear this model is now confused by longer sentences, sentences containing numerals and sentences that end with suspension points intuitively used here to illustrate that the text is an unfinished thought. For neural networks the length of input sentences is positively correlated with the error rate for classification tasks. This issue has been mitigated in recent years, but it still affects many deep learning architectures as the meaning of the sentence is progressively lost.

Additionally, all these specific categories of texts are under-represented in the training data and therefore it makes the predictability of an erroneous classification more plausible.

After increasing the learning rate and the number of training epochs for the fine-tuned altered XLM-RoBERTa variant it becomes increasingly apparent the problem lies with the long hateful texts. The model becomes more confused as the length of the text increases. It is the same phenomenon occurring with one variant of mBERT. Thus, a correlation can be made between these two powerful Transformer-based models and convolutional neural networks (CNNs), as they are both highly influenced by the length of their inputs and in some cases, this leads to erroneous decisions for classification.

Increasing the learning rate and the number of epochs even further leads to a decrease in the number of misclassified texts of 25% compared to the first time these two parameters were increased. Most of the errors made by this variant represent exclamatory Islamophobic sentences.

Surely there are multiple correlations between the errors made by these models that are not human-observable, however after a thorough visual inspection it is clear there is a high volume of misclassified hateful texts occurring not only in variants of the altered fine-tuned mBERT, but also in variants of XLM-RoBERTa. Therefore, to some extent many of these errors are not solely due to an architectural weakness in one model or another but are rather due in part to an architectural feature that applies to all those variants together.

## VI. RESULTS

The best performing model variant is a fine-tuned altered version of multi-lingual BERT trained for 10 epochs with a learning rate of 3e-4, which leads to a macro average of 0.67 for English/French classification. While no approach guarantees zero failure in the foreseeable future, false positives may continue triggering discussions of undesirable censorship [3] and that is something that should certainly be avoided, as the very reason machine learning methods for hate speech detection exist is to prevent the emergence of real chaotic censorship.

Fig. 1 and Fig. 2 present results corresponding exclusively to the model trained on English and tested on French, as intuitively this is the most important language pair of all four possible combinations, given the significantly higher volume of available English hate speech data compared to hate speech in other languages and thus assessing a real-world scenario of lucrative transfer learning.

Fig. 1 illustrates the results of the best combinations of parameter values for the number of epochs and learning rate. For mBERT, performance only increments slightly between the first two variants that are differentiated by an increase in learning rate and the number of training epochs. Increasing it further negatively impacts performance, by lowering the macro average score by 10%.

TABLE I.

| Model (EN-FR) | Epochs, Learning rates (macro avg.) | | |
|---|---|---|---|
| | *5 epochs, 1e-4* | *10 epochs, 3e-4* | *15 epochs, 5e-5* |
| mBERT | 0.66 | **0.67** | 0.57 |
| XLM-RoBERTa | 0.62 | 0.50 | 0.62 |

Fig. 1. Macro average results comparing all model variants (English to French)

Fig. 2 presents the results of both classifiers in terms of weighted average across different combinations of training parameters. Judging by this performance metric, the mBERT variant trained for 10 epochs with a learning rate of 3e-4 has achieved the best performance and once more it is closely followed by the variant trained for 10 epochs. Naturally the non-hateful texts correspond to a greater weight than the hateful ones, as these are more prevalent in the dataset.

TABLE II.

| Model (EN-FR) | Epochs, Learning rates (weighted avg.) | | |
|---|---|---|---|
| | *5 epochs, 1e-4* | *10 epochs, 3e-4* | *15 epochs, 5e-5* |
| mBERT | **0.73** | 0.72 | 0.59 |
| XLM-RoBERTa | 0.67 | 0.51 | 0.66 |

Fig. 2. Weighted average results comparing all model variants (English to French)

Both tables prove that by fine-tuning and altering multi-lingual BERT and XLM-RoBERTa for a hate speech detection task, BERT is superior not only in macro, but also in weighted average for most model variants presented here with different training hypermeters. A particularly interesting observation is the difference in weighted average for both models trained for 10 epochs with a learning rate of 3e-4, where the difference in the performance metric is 12%.

The results of the experiments performed to assess the performance of the variants of these models on English to English, French to French and French to English data are illustrated in Fig. 3 and Fig. 4. These figures are important because even though the most relevant language pair for this task in the real world is English/French, due to the considerably higher volume of available hate speech data in English compared to French, English-only, French-only, and French to English settings could prove insightful for how multi-lingual BERT and XLM-RoBERTa comprehend the meaning of hate by language.

TABLE III.

| Model | Extra language pairs (macro avg.) | | |
|---|---|---|---|
| | *EN-EN* | *FR-EN* | *FR-FR* |
| Best mBERT | **0.71** | 0.41 | **0.66** |

| Model | Extra language pairs (macro avg.) | | |
|---|---|---|---|
| | *EN-EN* | *FR-EN* | *FR-FR* |
| Best XLM-RoBERTa | 0.44 | **0.51** | 0.32 |

Fig. 3. Macro average results comparing best model variants on extra language pairs

Judging from the best model variants for each additional language pair, multi-lingual BERT is outperforming XLM-RoBERTa in two out of three of these cases. In singular language settings its macro average score is almost double, for both English-only and French-only data.

The results of the best fine-tuned altered XLM-RoBERTa admittedly are unfortunate for the monolingual setting, with scores less than 0.5, while mBERT did not transfer that well from French to English compared to all the other three combinations of language pairs.

TABLE IV.

| Model | Extra language pairs (weighted avg.) | | |
|---|---|---|---|
| | *EN-EN* | *FR-EN* | *FR-FR* |
| Best mBERT | **0.71** | 0.52 | **0.72** |
| Best XLM-RoBERTa | 0.43 | **0.55** | 0.27 |

Fig. 4. Weighted average results comparing best model variants on extra language pairs

The same situation occurs once more for weighted average scores, as it did for macro averages, where mBERT outperforms XLM-RoBERTa by far, thus reinforcing the intuition that multi-lingual BERT is the superior model in this case, at least given the range of values for the hyperparameters experimented with here.

Even when it comes to the weighted average as a performance metric, XLM-RoBERTa is not a suitable choice for this dataset at least when it comes to performing classification on a single language, especially in French with a score of less than 0.3, while mBERT manages to achieve over 0.5 while using this metric compared to the macro average.

The results of the presented work are satisfactory most likely since considering the whole sequence of inputs with Transformers models and using pre-trained word embeddings helps all models learn some common traits of hate speech. Still, hate speech is a difficult phenomenon to define and is not monolithic. This solution for detecting hateful texts, as is the case with many others, carries a significant amount of implicit bias within its training data.

## VII. CONCLUSIONS

Fine-tuned altered Transformer-based models serve their purpose faithfully when used to detect hate speech with cross-lingual (zero-shot) training, especially in our day and age when an increasing volume of such data stems from the realms of social media daily. It is true that Twitter and other Big Tech companies have revolutionized content publishing and communication, but it becomes more apparent every day that this is a double-edged sword [20].

Large sums of money are being spent both from the public and the private sector globally and yet the issue is hard to tackle, due to human nature itself, but also due to the very concept of the free Internet and the oftenly misunderstood right to free speech in a democracy. Changing behaviour through awareness is a possible alternative to regulation [6].

Up until a few years ago it was virtually impossible to accurately detect hate speech as this type of phrases come in highly varied forms and many of the previously employed supervised learning techniques could not realistically keep up with the textual subtleties, especially sarcasm and humor with which people infused their messages published on social media. Admittedly, the lack of reliable training data has contributed significantly to the delay of the day this task could be marked as a "solved problem." More insight is needed into understanding differences in hate speech across different languages [25].

The concept/mechanisms of "attention" has revamped the way natural language processing is done high-level. Conceptually, it has closed the gap even more between the way computers and humans think about language. Attention represents people's ability to focus only on certain sections of a phrase for example, in order to capture its meaning. Previous deep learning approaches for mainstream natural language processing tasks, whose state-of-the-art solutions now involve attention in one way or another, took the entire inputted text as it was then allowed a neural network to derive high-level features and encode these using word embeddings without enough consideration towards the hierarchy of importance of certain word groups over others.

Cross-lingual hate speech detection is a good approach to mitigate the lack of hate speech data in a certain language and most of the time that implies non-English data.

## VIII. FUTURE WORK

Capabilities of the classifiers presented here could be extended in several ways.

Firstly, such a classifier could be fine-tuned and have its performance assessed on other language pairs too other than any combination involving English/French, which would provide valuable insight into novel and efficient ways to tackle hate speech detection within less explored languages. There is a substantial amount of experimentation that is yet to be done also regarding training hypermeters and architectural variations. Alas, non-English data is desirable in this context, as the natural language processing industry is undoubtedly dominated by English data and in time this has inhibited the emergence of machine learning solutions trained on non-English datasets.

Additional features from tasks adjacent to hate speech detection such as sentiment analysis could prove highly useful as these would help understand the polarity of the language [11], however hate speech detection is more context-dependent than sentiment analysis and would always require high-level augmented features, including word representations, for good deep learning performance [1].

Given more time, data containing hate speech belonging to other languages should be mined more extensively and incorporated into datasets for hate speech classifiers so that deep learning approaches with native training in a certain language could emerge. These could perform better than many cross-lingual approaches from English to another language. Efficient deep learning solutions imply a large volume of relevant training data collected with a well-documented consistent methodology.

Secondly, an end-to-end hate speech detection system could be designed and developed within a robust cloud ecosystem (GCP, AWS or Azure) and make use of its unique capabilities. The option of integrating load balancers that would enable APIs to be highly available in case of a spike in usage and the option to use reliable cloud data storage solutions such as data lakes to store impressive volumes of historical data with the purpose of analyzing it later and consequently improve future models using previously derived insights from that data. Such a system could be used to filter tweets before content recommendations or be integrated within a higher level chatbot that would be able to comprehend the true meaning and intent of hate speech and act to it accordingly.

Lastly, within a cloud environment the barrier set by limited computational resources could be seamlessly lifted to be able to perform an extensive set of configurations for a given deep learning architecture [6].

REFERENCES

[1] H. Watanabe et B. Bouazizi, T. Ohtsuki, "Hate speech on Twitter: A pragmatic approach to collect hateful and offensive expressions and perform hate speech detection," Yokohama, Japan, 2018.

[2] S. Biere, S. Bhulai, "Hate speech detection using natural language processing techniques," Amsterdam, Netherlands, 2018.

[3] S. Koffer, D. M. Riehle, S. Hogenberger, J. Becker, "Discussing the value of automatic hate speech detection in online events," Munster, Germany, 2018.

[4] M. Mozafari, R. Farahbakhsh, N. Crespi, "A BERT-based transfer learning approach for hate speech detection in online social media," Paris, France, 2019.

[5] S. MacAvaney et al., "Hate speech detection: Challenges and solutions," Washington DC, United States of America, 2019.

[6] S. Zimmerman, C. Fox, U. Kruschwitz, "Improving hate speech detection with deep learning ensembles," Essex, United Kingdom, 2018.

[7] Vaswani et al., "Attention is all you need," arXiv:1706.03762, 2017.

[8] S. Benesch, "Dangerous speech: A proposal to prevent group violence," 2012.

[9] E. Ombui, L. Muchemi, P. Wagacha, "Hate speech detection in code-switched text messages," Nairobi, Kenya, 2019.

[10] "Kenya to monitor social media during elections", The EastAfrican, 12-Jan-2017.

[11] G. Koushik, K. Rajeswari, S. K. Muthusamy, "Automated Hate Speech Detection on Twitter," Punc, India, 2019.

[12] U. R. Hodeghatta, "Sentiment Analysis of Hollywood Movies on Twitter," Proceedings of the 2013 IEEE/ACM International Conference on Advances in Social Networks Analysis and Mining," ASONAM 13, 2013.

[13] S. Hinduja, J.W. Patchin, "Bullying. Cyberbullying and Suicide," Archives of Suicide Research Cyberbullying and Suicide, vol. 14, no. 3, 2010, pp. 206-221.

[14] C. Nobata, J. Tetreault, A. Thomas, Y. Mehdad, Y. Chang, "Abusive Language Detection in Online User Content," International World Wide Web Conference Committee (IW3C2), Montreal, Canada, 2016.

[15] P. Badjatiya, S. Gupta, M. Gupta, V. Varma, "Deep Learning for Hate Speech Detection in Tweets," Hyderabad, India, 2017.

[16] J. Lee, "Twitter apologizes for mishandling reported threat from mail-bomb suspect," 2018.

[17] A. Schmidt, M. Wiegand, "A survey on hate speech detection using natural language processing," Proceedings of the Fifth International Workshop on Natural Language Processing for Social Media, 2017, pp. 1-10.

[18] R. D. King, G. M. Sutton, "High times for hate crimes: Explaining the temporal clustering of hate-motivated offending," Criminology, vol. 51, no. 4, 2013, pp. 871-894.

[19] Ross et al., "Measuring the reliability of hate speech annotations: The case of the European refugee crisis," preprint arXiv:1701.08118, 2017.

[20] Z. Zhang, L. Luo, "Hate Speech Detection: A Solved Problem? The Challenging Case of Long Tail of Twitter," arXiv:1803.03662v2, 2018.

[21] I. Galiardone, D. Gal, T. Alves, G. Martinez, "Countering online hate speech," UNESCO Series on Intternet Freedom, 2015, pp. 1-73.

[22] B. Gamback, U. K. Sikdar, "Using convolutional neural networks to classify hate speech" Proceedings of the First Workshop on Abusive Language Online, pp. 85-90, Association for Computational Linguistics, 2017, doi:10.18653/v1/W17-3013.

[23] Y. Cheng, Y. Zhou, S. Zhu, H. Xu, "Detecting offensive language in social media to protect adolescent online safety," Proceedings of the 2012 SE/IEEE International Conference on Privacy, Security, Risk and Trust, SOCIALCOM-PASSAT 2012, pp. 71-80, Washington, DC, USA, 2012, IEEE Computer Society, doi:10.1109/SocialCom-PASSAT.2012.55.

[24] Z. Waseem, D. Hovy, "Hateful symbols or hateful people? Predictive features for hate speech detection on Twitter," Proceedings of the NAACL Student Research Workshop, pp. 88-93, Association for Computational Linguistics, 2016, doi:10.18653/v1/N16-2013.

[25] Corazza et al., "Comparing Different Supervised Approaches to Hate Speech Detection," 2018.

[26] T. Davidson, D. Bhattacharya, I. Weber, "Racial Bias in Hate Speech and Abusive Language Detection Datasets," 2017.

[27] A. Arango, J. Perez, B. Poblete, "Hate speech detection is not as easy as you may think: A closer look at model validation (extended version)," Santiago, Chile, 2020.

[28] N. Ousidhoum, Z. Lin, H. Zhang, Y. Song, D. Yeung, "Multilingual and Multi-Aspect Hate Speech Analysis," arXiv: 1908.11049v1, 2019.

[29] Y. Chung, E. Kuzmenko, S. S. Tekiroglu, M. Guerini, "CONAN - Counter Narratives through Nichesourcing: A Multilingual Dataset of Responses to Fight Online Hate Speech," Proceedings of the 57th Annual Meetings of the Association for Computational Linguisitics, pp. 2819-2829, 2019, doi:10.18653/v1/P19-1271.

[30] J. P. Breckheimer, "A haven for hate: The foreign and domestic implications of protecting Internet hate speech under the first amendment," South California Law Rev., vol. 75 no. 6, p. 1493, 2002.